\documentclass[10pt,twocolumn,letterpaper]{article}

\usepackage{cvpr}
\usepackage{times}
\usepackage{graphicx}
\usepackage{amsmath}
\usepackage{amssymb}
\usepackage{booktabs}
\usepackage{graphics} 
\usepackage{epsfig} 
\usepackage{bbm}
\usepackage{commath}
\DeclareMathOperator*{\argmax}{arg\,max}

\usepackage{float}
\usepackage{cite}
\usepackage{multirow}
\usepackage{svg}
\usepackage{algorithm}
\usepackage{algpseudocode}
\usepackage{import}
\usepackage{subcaption}

\usepackage[pagebackref=true,breaklinks=true,letterpaper=true,colorlinks,bookmarks=false]{hyperref}

\usepackage[capitalize]{cleveref}
\crefname{section}{Sec.}{Secs.}
\Crefname{section}{Section}{Sections}
\Crefname{table}{Table}{Tables}
\crefname{table}{Tab.}{Tabs.}

\begin{document}

\title{Active Learning for Object Detection with Non-Redundant Informative Sampling}

\author{Aral Hekimoglu\\
Technical University Munich\\
Munich, Germany\\
{\tt\small aral.hekimoglu@tum.de}
\and
Adrian Brucker\\
Technical University Munich\\
Munich, Germany\\
{\tt\small adrian.brucker@tum.de}
\and
Alper Kagan Kayali\\
Technical University Munich\\
Munich, Germany\\
{\tt\small alper.kagan.kayali@tum.de}
\and
Michael Schmidt\\
BMW Group\\
Munich, Germany\\
{\tt\small michael.se.schmidt@bmw.de}
\and
Alvaro Marcos-Ramiro\\
BMW Group\\
Munich, Germany\\
{\tt\small alvaro.marcos-ramiro@bmw.de}
}
\maketitle

\begin{abstract}

Curating an informative and representative dataset is essential for enhancing the performance of 2D object detectors. We present a novel active learning sampling strategy that addresses both the informativeness and diversity of the selections. Our strategy integrates uncertainty and diversity-based selection principles into a joint selection objective by measuring the collective information score of the selected samples. Specifically, our proposed NORIS algorithm quantifies the impact of training with a sample on the informativeness of other similar samples. By exclusively selecting samples that are simultaneously informative and distant from other highly informative samples, we effectively avoid redundancy while maintaining a high level of informativeness. Moreover, instead of utilizing whole image features to calculate distances between samples, we leverage features extracted from detected object regions within images to define object features. This allows us to construct a dataset encompassing diverse object types, shapes, and angles. Extensive experiments on object detection and image classification tasks demonstrate the effectiveness of our strategy over the state-of-the-art baselines. Specifically, our selection strategy achieves a 20\% and 30\% reduction in labeling costs compared to random selection for PASCAL-VOC and KITTI, respectively.

\end{abstract}

\section{Introduction} \label{sec:introduction}

\begin{figure}[t]
    \centering
    \includegraphics[width=1.0\linewidth]{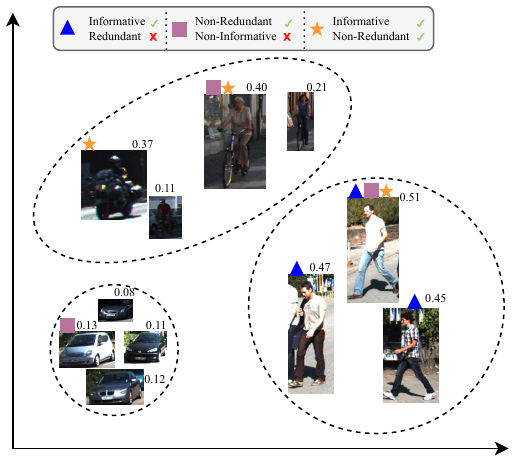}
    \caption{Illustration of different AL selection strategies in feature space. The size of each object symbolizes its uncertainty score, also denoted next to the object. Dashed lines mark clusters formed by k-means. The uncertainty-based method \cite{yoo2019learning} (triangle) selects objects with the highest uncertainty. However, due to the similarity among the selected objects, there is redundancy in the selections. The diversity-based method \cite{sener2018active} (square) selects one object from each cluster. However, since the selected \textit{Car} belongs to a well-represented class and has low uncertainty, its selection is less informative. Our proposed approach, NORIS (star) selects object with high uncertainty while maintaining diversity.}
    \label{fig:intro_overview}
\end{figure}

Accurately detecting 2D objects \cite{liu2022swin, xu2021end, duan2019centernet} is fundamental in scene understanding across various applications, such as autonomous driving. The current state-of-the-art (SOTA) object detection systems rely on deep learning techniques and heavily depend on large-scale annotated datasets. However, annotating extensive amounts of 2D detection data is time-consuming and labor-intensive, making labeling the entirety of the collected data challenging. Moreover, selecting similar samples might lead to overfitting, where the model becomes overly specialized in some areas of the input space, limiting its generalization ability. Therefore, curating an informative dataset that is also representative of the target distribution is crucial to achieving optimal object detection performance.

Active learning (AL) \cite{elezi2022not, sener2018active, gal2017deep} emerged as a promising solution for selecting the most suitable data for labeling. Uncertainty-based AL methods \cite{roy2018deep, desai2019adaptive, tang2021towards, li2021deep, choi2021active, hekimoglu2022efficient} leverage the uncertainty of the output of the network to assign an information score to each sample and subsequently select samples with the highest score for labeling. However, as depicted in \cref{fig:intro_overview}, these methods tend to select highly similar samples, such as the three pedestrians with similar poses, leading to redundancy and a lack of diversity within the labeled dataset.

To prevent redundancy in selections, diversity-based methods \cite{sener2018active, sinha2019variational, shen2019kcovers, agarwal2020contextual} select representative samples that cover the input space. One approach is to cluster the data and choose one sample from each cluster. However, selecting the most diverse samples does not necessarily guarantee informativeness. As shown in \cref{fig:intro_overview}, the diversity-based method selects a \textit{Car} because it belongs to a distinct cluster from the other selected samples. However, this selection is less informative as the car has a lower uncertainty compared to other samples.

To address the limitations of the aforementioned methods, hybrid approaches \cite{kirsch2019batch, cao2021bayesian, tan2019batch, sharma2019active} integrate uncertainty and diversity by multiplying the individual uncertainty and diversity scores. However, these methods treat uncertainty and diversity as separate properties and do not fully capture the interaction between sample uncertainties. For instance, in \cref{fig:intro_overview}, learning from one of the pedestrians would decrease the uncertainty of nearby pedestrians. To better estimate the collective information score of a selected set, modeling the interaction of uncertainties based on the distance of samples is necessary.

We propose a novel sampling strategy called NOn-Redundant and Informative Sampling (NORIS) in \cref{sec:method-redundancy}. NORIS tackles the challenges of redundancy and informativeness by formulating the active learning objective as a collective information score. Our work is the first to formally integrate the interaction (as defined by the diversity) of different uncertainties within the same AL selection objective and offer an effective algorithmic solution.

Furthermore, previous AL methods for object detection primarily rely on the distance between image features to define diversity. As a result, these methods select a diverse range of scenes, including variations in lighting and weather conditions. However, their holistic image approach makes it difficult to capture the distinct diversity of objects within these scenes. To overcome this limitation, in \cref{sec:method-diversity}, we propose to calculate the distance using object features, which we define as the intermediate network features within the region of interest (ROI) for each detected box. This allows us to capture the diversity of objects within a scene. With our approach, the detector is trained on a more comprehensive and diverse set of object instances, enabling it to generalize better and achieve improved detection performance across different scenarios.

Our main contributions are summarized as follows:
\begin{itemize}
    \item We propose NORIS, a novel strategy that addresses the challenge of selecting informative samples while minimizing redundancy by formulating the AL selection objective as a collective information score.
    \item We introduce a unique approach to define diversity by utilizing object features extracted from the detector, enhancing the generalizability of the detector to a diverse set of objects.
    \item We conduct an extensive experimental evaluation to assess the performance of NORIS against the existing SOTA AL baselines, demonstrating superior performance and up to 30\% reduction in labeling costs.
\end{itemize}

\section{Related work} \label{sec:related}

\subsection{Uncertainty-based methods} \label{sec:related_uncertainty}

Most AL methods for object detection rely on an information score, such as the uncertainty of the model in its predictions, to prioritize the selection of the most informative samples for labeling. For object detection, uncertainty-based methods typically target either the classification uncertainty or the localization uncertainty of the predicted bounding box. Classification uncertainty-focused methods utilize the predicted class probability distribution as an estimate of the model's confidence for each sample \cite{roy2018deep, brust2018active, aghdam2019active, haussmann2020scalable, yuan2021multiple}. Localization uncertainty methods \cite{schmidt2020advanced, kao2018localization, choi2021active} use the uncertainty associated with the bounding box parameters such as position or dimensions to formulate an information score. Kao \etal \cite{kao2018localization} introduced two uncertainty measures, namely localization tightness and localization consistency, for sample selection. Choi \etal \cite{choi2021active} proposed an aleatoric uncertainty head designed to estimate localization uncertainty.

Another approach to quantify informativeness involves measuring the disagreement among different predictions for the same input. Epistemic uncertainty-based methods leverage diverse predictions obtained from multiple models, where each model is trained with different initial random weights \cite{chitta2020training}, or from a single model with a dropout layer that generates distinct predictions at each forward pass \cite{miller2018dropout}. In inconsistency-based methods \cite{elezi2022not, yu2022consistency}, diverse predictions are obtained by applying various augmentations to the same input sample.

A task-agnostic AL approach, LL4AL, was proposed by Yoo \etal \cite{yoo2019learning}. They include an additional head in the network designed to predict the target loss for unlabeled inputs, and the information score is defined based on this predicted loss.

\subsection{Diversity-based active learning}\label{sec:related_diversity}

A common issue with uncertainty-based methods is the tendency to select similar samples, leading to redundant selections. In contrast, diversity-based methods \cite{sener2018active, sinha2019variational, ebrahimi2020minimax, wang2018uncertainty} select representative samples that cover the entire dataset space. These methods apply a similarity measure between a candidate and previously selected samples, choosing the sample with the least similarity to the previous selections. Diversity-based methods can be categorized into three groups based on the type of similarity measurement: feature-based \cite{sener2018active}, adversarial \cite{sinha2019variational, ebrahimi2020minimax}, and context-based \cite{agarwal2020contextual}.

Feature-based methods represent each sample with a feature vector and employ a distance metric to estimate the similarity. Sener \etal \cite{sener2018active} proposed constructing a core-set by efficiently solving the k-center problem, where distances are computed as the Euclidean norm between features extracted from intermediate layers of a network for each image. Adversarial diversity-based methods \cite{sinha2019variational} train a discriminator to predict whether a sample belongs to the previously selected set. The output from the discriminator is then used as the similarity measure to the labeled subset. Agarwal \etal. \cite{agarwal2020contextual} introduced a method for contextual diversity, in which the KL-divergence between the predicted class probabilities of a sample and the previously selected subset is used to define contextual similarity.

Existing diversity-based methods primarily focus on images and have not been adapted to object detection. Haussmann \etal \cite{haussmann2020scalable} implemented a feature-based diversity method for object detection but relied on image features. In our work, we propose leveraging object features extracted from intermediate layers of an object detector, aiming to construct a training set with a more diverse set of objects.

\subsection{Hybrid uncertainty-diversity methods}\label{sec:related_hybrid}

Considering both uncertainty and diversity is critical in forming an effective selection strategy that identifies informative samples while simultaneously avoiding redundancy \cite{kirsch2019batch, cao2021bayesian, tan2019batch, sharma2019active, shen2019kcovers, zhdanov2019diverse, ash2019deep}. Shen \etal \cite{shen2019kcovers} proposed K-Covers, a method that first clusters the unlabeled set using a modified k-means algorithm and then selects the most uncertain sample from each cluster. Zhdanov \etal \cite{zhdanov2019diverse} proposed DBAL, which leverages a weighted k-means clustering algorithm by multiplying the distance by the uncertainty to prioritize both diversity and informativeness of a sample. Haussmann \etal \cite{haussmann2020scalable} adopt a similar approach by multiplying the distance between feature vectors with the uncertainty and selecting samples based on the resultant scores. Ash \etal introduced BADGE \cite{ash2019deep}, which uses gradient embeddings as a proxy for uncertainty and iteratively selects a batch of samples that are both diverse and uncertain based on their distances in the gradient embedding space.

In contrast to prior hybrid methods that derive a singular selection score by multiplying the uncertainty and diversity of a sample, our approach focuses on the information score of a set of samples considered collectively. We propose a unique approach to integrate uncertainty and diversity by assessing the impact that labeling and training a sample would have on the information score of other samples, with the impact scaled based on their similarities.

\section{Methodology} \label{sec:methods}

\subsection{Problem definition} \label{sec:method-problem}

Let $(x,y)$ denote a sample pair drawn from a labeled training dataset $(X_{\text{train}},Y_{\text{train}})$ where $X_{\text{train}}$ represents the set of labeled data points and $Y_{\text{train}}$ their corresponding labels. Let $u$ denote an unlabeled sample drawn from a larger pool of unlabeled samples $X_{\text{U}}$. In each AL cycle, a model $\phi$ is trained with the labeled data $(X_{\text{train}},Y_{\text{train}})$. Then, an acquisition function selects a subset of unlabeled samples to be manually labeled by an external oracle. These newly labeled samples are then added to the labeled set for training in the next iteration. We denote the selected subset by $S\subseteq X_{\text{U}}$, and $B=\abs{S}$ represents the labeling budget in each iteration. Consequently, the AL objective can be defined as selecting a subset $S$ of unlabeled data points such that the performance of the model trained on $(X_{\text{train}} \cup S, Y_{\text{train}} \cup Y_{\text{S}})$ is maximized, where $Y_{\text{S}}$ represents the labels for the samples in $S$.

\subsection{Non-redundant informative sampling} \label{sec:method-redundancy}

NORIS encapsulates proposed query strategies that combine uncertainty with diversity by incorporating redundancy. We define a function $\sigma: X \to \mathbbm{R}_{\geq 0}$ that assigns each unlabeled data $u \in X_{\text{U}}$ an information score $\sigma(u)$ (\eg, the model’s uncertainty for that image). In the setting of uncertainty-based AL, in each query iteration, the $B$ highest-ranked samples are selected for annotation. This can be formulated as maximizing the sum of their information scores:
\begin{equation}
\label{eq:objective}
\argmax_{S \subseteq X_{\text{U}},\ \abs{S} = B} \sum_{u \in S} \sigma(u)
\end{equation}

One drawback of this approach is that the information gain from a sample is reduced if we already select and train with a similar, more informative sample.

To model this interaction, we consider the loss of a sample as our selection criteria. Training on one sample reduces the loss of similar samples as the network learns to generalize in that region in the optimization space. Formally, we follow the proof in Core-Set \cite{sener2018active} and define the loss function to be $\kappa$-Lipschitz continuous. With this definition, we express the upper bound on the loss of one sample $u$ after training with another sample $v$ (details provided in the supplementary material):
\begin{equation}
\label{eq:loss_bound}
    l(u; \theta_{\text{i+1}}) \leq l(u; \theta_{\text{i}}) + 2 \cdot \kappa \cdot d(u, v) - l(v; \theta_{\text{i}})
\end{equation}

where $d(u,v)$ represents the distance between $u$ and $v$, and $\theta_{\text{i}}$ and $\theta_{\text{i+1}}$ denote the parameters of the network before and after a training iteration that includes sample $v$. Following this argumentation, the final loss of a sample $u$ is bounded by an expression inversely related to the initial loss of sample $v$ and directly related to the distance between samples $u$ and $v$. This implies that closer samples have a stronger influence on reducing each other's loss, while higher initial loss of the trained sample $v$ leads to a tighter bound in the loss of the neighboring sample $u$.

Therefore, to get an information score of a set collectively, we need to model the interaction of training on one sample would have on the information score of other samples. We propose a model to update the information score of the sample u, assuming that v is selected. 
\begin{equation}
\label{eq:update_sigma}
    \sigma^{'}(u):= \sigma(u) - sim(u,v) * \sigma(v)
\end{equation}

Here, we define a function that measures the similarity between samples sim$(u,v)$ to be inversely correlated to $d(u,v)$. We require that sim$(u, u) = 1$ for all $u \in X_{\text{U}}$ and that sim is symmetric, i.e. sim$(u, v) =$ sim$(v, u)$ for all $u, v \in X_{\text{U}}$. The more similar $u$ and $v$ are, the higher sim$(u, v)$ should be. We propose two similarity measures that utilize the distance metric $d$:
\begin{align}
\label{eq:sim-gaussian}
    \text{Gaussian: sim}(u, v) & = e^{- \frac{1}{\lambda} d(u, v)^2}\\
\label{eq:sim-linear}
    \text{linear: sim}(u,v) & = \max\left\{0, 1 - \frac{d(u, v)}{\lambda}\right\}
\end{align}

The hyperparameter $\lambda > 0$ indicates the influence of a sample to its surroundings. Higher values represent stronger influence, while lower values lead to weaker influence. An exhaustive hyperparameter search can assist in finding an appropriate value. For the linear similarity, we suggest experimenting with values in the range of $(0, d_{\text{max}}]$, where $d_{\text{max}}$ is defined as:
\begin{equation}
    d_{\text{max}} := \max_{x, x^\prime \in X_U} d(x, x^\prime).
\end{equation}

We argue that values for $\lambda$ smaller than or equal to the maximum distance between the unlabeled samples are reasonable, as the similarity score of the most distant samples in the embedding space should be zero. Note that $d_{\text{max}}$ must be recalculated in each AL cycle since distances are calculated between the changing low-dimensional embeddings. This also allows for automatic adaptation to the scale of the feature space as opposed to fixing a value throughout all AL iterations. When conducting a hyperparameter search for the linear similarity, we fix a value $\alpha \in (0, 1]$ and scale it to $\lambda = \alpha \cdot d_{\text{max}}$ in each AL cycle. For the Gaussian similarity, we fix a value $\alpha \in (0, 1]$ and scale it to $\lambda = \frac{(\alpha \cdot d_{\text{max}})^2}{\pi}$ in each AL cycle for hyperparameter tuning. For additional details, please refer to the supplementary material.

We propose two variants on how to incorporate \cref{eq:update_sigma} into an AL selection objective called NORIS-Sum and NORIS-Max. Both versions are not restricted to a specific method of measuring the information value of individual images. The choice of similarity measure between images is also flexible, making NORIS highly versatile.

\textbf{NORIS-Sum.} In NORIS-Sum, we define the AL selection objective as the aggregate of the information scores. We determine the individual score of a sample $u$ as the information score of $u$, deducted by the weighted sum of information scores of all the selected samples $v \in S \setminus \{u\}$, scaled by their similarity $\text{sim}(u, v)$. We sum the individual scores to generate a batch-wise acquisition score that better captures the true value of a set of samples.
\begin{equation}
\label{eq:noris_sum}
    \argmax_{S \subseteq X_{\text{U}},\ \abs{S} = B} \sum_{u \in S} \sigma(u) - \left(\sum_{v \in S \setminus {u}} \text{sim}(u, v) * \sigma(v) \right)
\end{equation}

To efficiently solve this objective, we propose the NORIS-Sum algorithm (\cref{alg:noris-sum}). In each iteration of the while-loop, the sample $u_{next}$ that leads to the highest immediate gain is selected and added to $S$. Subsequently, for each $u \in X_{\text{U}} \setminus S$, the information score is updated by subtracting the uncertainty of $u_{\text{next}}$ scaled by the similarity $\text{sim}(u, u_{\text{next}})$. This algorithm has a runtime complexity of $\mathcal{O}(B \cdot \lvert X_{\text{U}} \rvert)$.
\begin{algorithm}
    \caption{NORIS-Sum}
    \label{alg:noris-sum}
    \begin{algorithmic}
        \Require unlabeled dataset $X_{\text{U}}$, batch size $1 \leq B \leq \abs{X_{\text{U}}}$ 
        \State $S \gets \emptyset$
        \While{$\abs{S} \neq B$}
\State $u_{\text{next}} \gets \argmax\limits_{u \in X_{\text{U}} \setminus S} \sigma(u)$
            \State $S \gets S \cup \{u_{\text{next}}\}$
            \For{$u \in X_{\text{U}} \setminus S$}
                \State $\sigma(u) \gets \sigma(u) - \text{sim}(u, u_{next}) * \sigma(u_{next})$
            \EndFor
        \EndWhile
    \end{algorithmic}
\end{algorithm}

\begin{figure*}[t]
    \centering
    \includegraphics[width=0.90\linewidth]{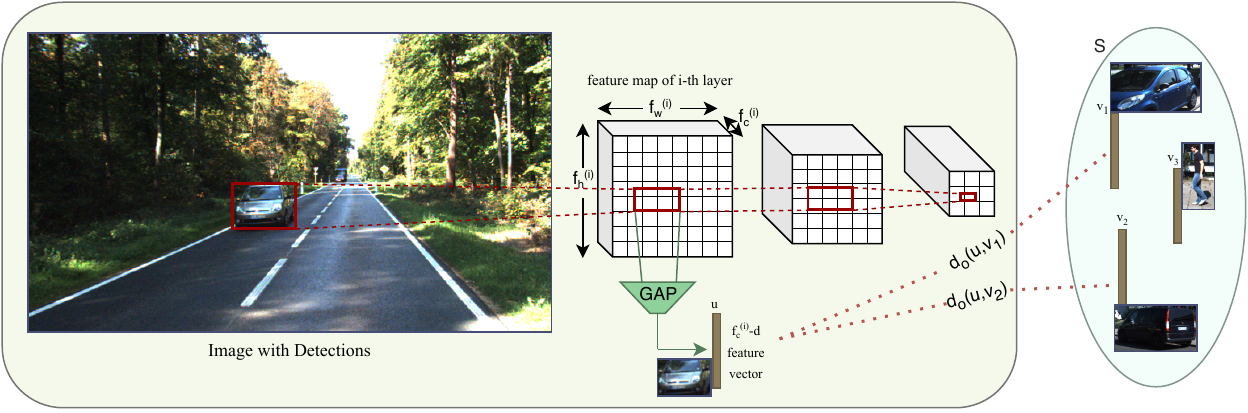}
    \caption{Illustration of the object-diversity calculation. For each detected object, we crop the corresponding ROI from intermediate feature maps and apply global average pooling to get a feature vector. We then measure the distance between the feature vector of the detected object and the feature vectors of objects in selected set $S$ to define the distance metric.}
    \label{fig:overview}
\end{figure*}

\textbf{NORIS-Max.} In NORIS-Max, instead of reducing the information score of an individual image for each sample in its vicinity, we consider only the impact of the closest one. We argue that distant data has less influence, as the loss bound in \cref{eq:loss_bound} becomes less tight as the distance increases. Accumulating many small similarity scores can have a strong impact which might not be desirable. Therefore, in this approach, the score of a sample is reduced only by the similarity of its closest sample in the batch of selected points. The batch-wise acquisition function becomes:
\begin{equation}
\label{eq:noris-max}
    \argmax_{S \subseteq X_{\text{U}},\ \abs{S} = B} \sum_{u \in S} \sigma(u) - \text{sim}(u,v_{\text{c}}) \cdot \sigma(v_{\text{c}})
\end{equation}

where $v_{\text{c}}$ represents the most similar sample to sample $u$. We follow the same algorithm described in \cref{alg:noris-sum}, and modify the update step to only consider its closest sample.

\subsection{Object diversity} \label{sec:method-diversity}

One of the critical design choices in our algorithm is the distance metric $d(u,v)$. While the Euclidean distance between intermediate features of a deep neural network is a common choice for image classification \cite{sener2018active}, we propose a novel approach that uses features describing the object instead of features for the whole image. This approach is more intuitive and aligns better with the goal of covering a diverse set of objects with different variations in types, angle, and occlusion patterns to make the object detector more generalizable.

Our approach, illustrated in \cref{fig:overview}, leverages object features to compute the similarity between objects. We use our object detector to detect all objects in the unlabeled pool and extract intermediate feature maps $f^{(i)}$ with dimensions of ($f_w^{(i)},f_h^{(i)},f_c^{(i)}$). We compute the ROI for each detected object in the feature map $f^{(i)}$, defined by the its bounding box coordinates. We scale the location $(o_x*f_w^{(i)}/ i_w, o_y * f_h^{(i)}/ i_h)$ and size $(o_w*f_w^{(i)}/ i_w, o_h * f_h^{(i)} / i_h)$, where $(o_x, o_y, o_w, o_h)$ represent the pixel-wise location, width, and height of the detected object, and $(i_w, i_h)$ represent the width and height of the input image. Next, we crop the corresponding region from the feature map, apply global average pooling (GAP), and obtain a feature vector with $f_c^{(i)}$ elements. These features are then used to compute the Euclidean distance between objects.

To compute the distance between images based on the objects they contain, we aggregate the distances between the objects by taking the maximum distance between any two objects in the images. Thus, for two images $u$ and $v$, the object-based distance $d_o(u,v)$ is defined as follows:
\begin{equation}
d_o(u,v) = \max_{o_u \in u, o_v \in v} |f_{o_u}-f_{o_v}|^2
\end{equation}

Here, $o_u$ and $o_v$ represent the sets of detected objects in the images $u$ and $v$, respectively, and $f_{o_u}$ and $f_{o_v}$ represent the feature vectors for each object. The maximum distance is selected since the overall distance between images should reflect the distance between the most dissimilar objects in the images.

In addition to object features, we incorporate scene diversity into our method to capture a diverse set of scenes with varying weather and lighting conditions and to be able to measure the distance between two images in the absence of detected objects. To obtain image features, we adopt a similar approach as with object features by applying global average pooling with ROI defined as the entire image. For any two images $u$ and $v$, we define the final distance $d(u,v)$ as:
\begin{equation}
d(u,v) = d_o(u,v) * |f_{u}-f_{v}|^2
\end{equation}

Here, $f_{u}$ and $f_{v}$ are the global image features for images $u$ and $v$, respectively. This definition of distance captures both object and scene diversity, allowing our algorithm to select samples that cover a diverse range of object variations and scenes.

\section{Experiments} \label{sec:experiments}

\begin{figure*}
  \centering
  \begin{subfigure}{0.45\linewidth}
    \includegraphics[width=1.0\linewidth]{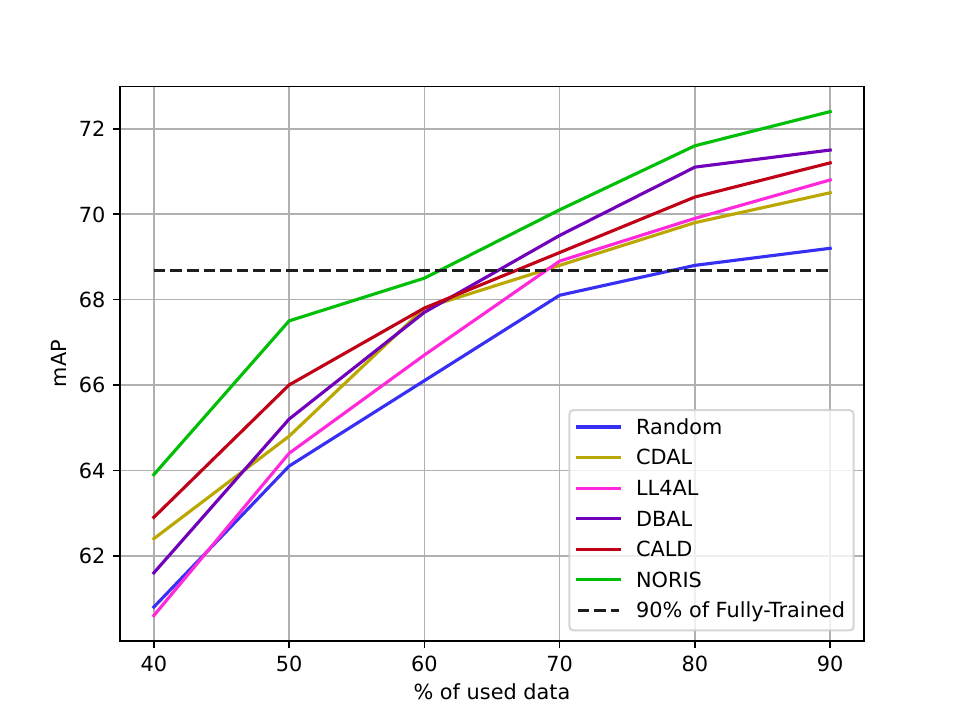}
    \caption{PASCAL-VOC 2007}
    \label{fig:pascal}
  \end{subfigure}
  \hfill
  \begin{subfigure}{0.45\linewidth}
    \includegraphics[width=1.0\linewidth]{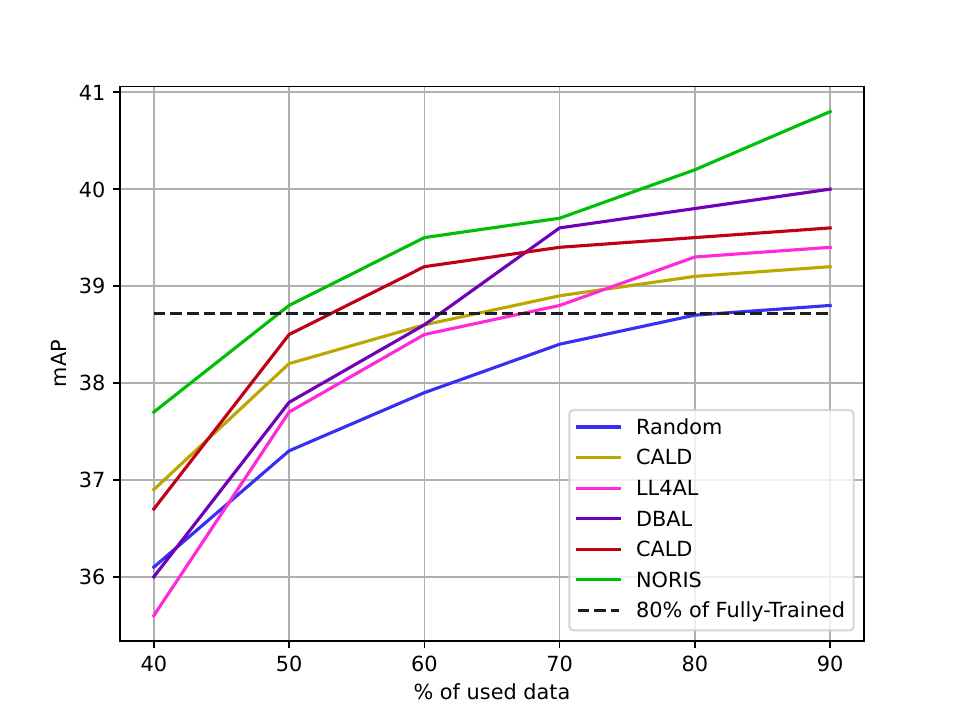}
    \caption{KITTI \textit{val}}
    \label{fig:kitti}
  \end{subfigure}
  \begin{subfigure}{0.45\linewidth}
    \includegraphics[width=1.0\linewidth]{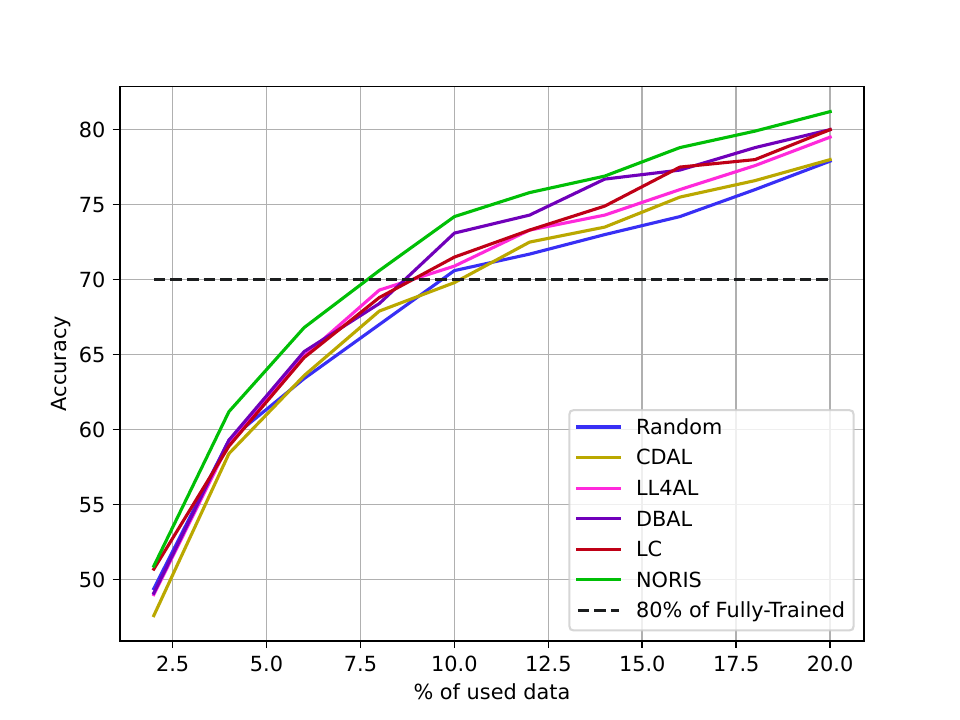}
    \caption{CIFAR 10}
    \label{fig:cifar-10}
  \end{subfigure}
  \hfill
  \begin{subfigure}{0.45\linewidth}
    \includegraphics[width=1.0\linewidth]{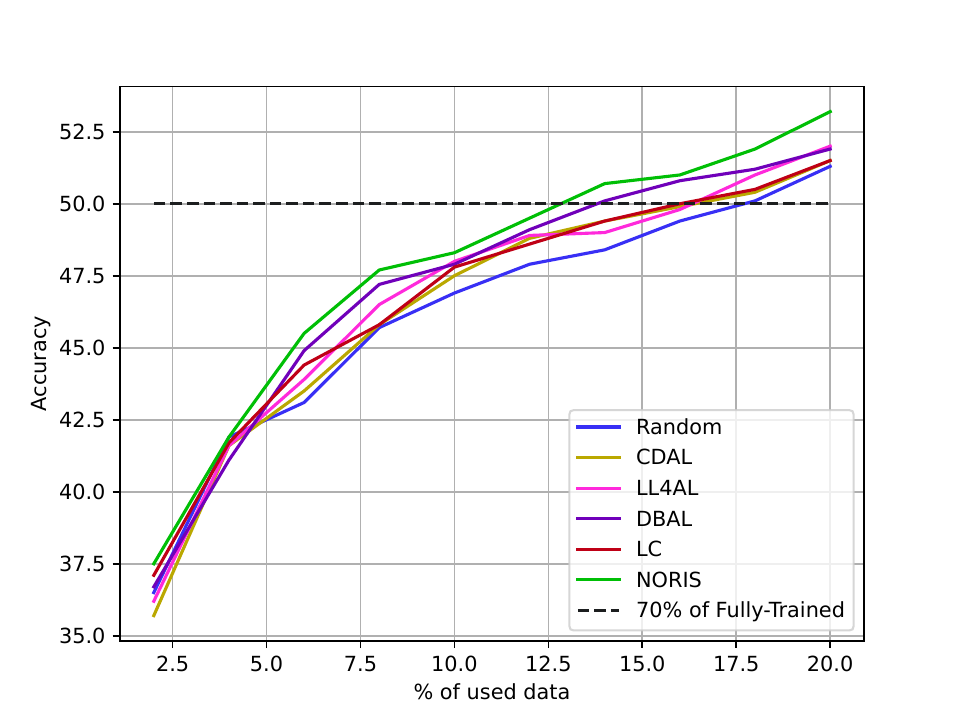}
    \caption{CIFAR 100}
    \label{fig:cifar-100}
  \end{subfigure}
  \caption{Comparison with SOTA AL methods. Lines indicate the averaged results over three trials. Note that all methods start from the same network trained with the initially labeled data.}
  \label{fig:alod}
\end{figure*}

\subsection{Comparison with baselines}\label{sec:exp_main}

\textbf{Datasets and evaluation metric.} We evaluate the performance of our proposed algorithm on four datasets, two for object detection: PASCAL VOC \cite{everingham2010pascal}, and KITTI \cite{geiger2012we}, and two for image classification: CIFAR-10 \cite{krizhevsky2009learning}, and CIFAR-100 \cite{krizhevsky2009learning}. We use the mean Average Precision (mAP)\cite{feng2021review} as the evaluation metric for object detection and accuracy for image classification. For PASCAL VOC, we combine the 2007 and 2012 datasets, resulting in 16,551 samples, which serve as the unlabeled pool. We evaluate the performance on the test set of PASCAL VOC 2007. For KITTI, we follow the data split used in \cite{lang2019pointpillars}. The training set consists of 7,481 images, which are divided into an unlabeled pool with 3,712 samples and a validation set with 3,769 samples. We report the results on the validation set. For the CIFAR-10 and CIFAR-100 datasets, we use the training set containing 50,000 images as the initial unlabeled pool and report results on the test set.

\textbf{Model architecture.} We use CenterNet \cite{zhou2019objects} as our base detector. We train the model for the same number of epochs and follow the same hyperparameters and optimization scheme described in the original paper. The experiments are conducted on an NVIDIA Tesla V100 GPU using the PyTorch deep learning framework \cite{paszke2019pytorch}. For the image classification experiments, we use ResNet-18 \cite{he2016deep} as our base classifier and extract the image feature embeddings from the penultimate layer to calculate distances. In each query iteration, we train the model continuously for 50 epochs with an early stopping patience of 10, using the Adam optimizer with a learning rate of $10^{-3}$.

\textbf{Active learning details.} We start our experiments with an unlabeled pool of samples and create an initial training set by randomly selecting samples. We randomly split the unlabeled pool into a labeled pool, which serves as the initial training set. At each AL iteration, we select samples from the remaining unlabeled pool to add to the training set using the selection strategy. To simulate the labeling process, we use the already available annotations. For the object detection datasets, we initially select 30\% of the available samples and add 10\% more during each AL cycle. For image classification tasks, we initially select 1000 images (equivalent to 2\% of the total dataset) and add 1000 images in each iteration. We report the mean of the respective metric for each experiment over three independent runs with different random selections of the initial labeled pool. We also report the variances across these three runs in the supplementary material.

\textbf{Baselines.} To evaluate the effectiveness of our proposed selection strategy, we compare it against several baselines from the literature. We choose two SOTA uncertainty-based methods: the inconsistency-based selection strategy \textbf{CALD} \cite{yu2022consistency}, and the loss-based method \textbf{LL4AL} \cite{yoo2019learning}. Additionally, we compare against the diversity-based method \textbf{CDAL} \cite{agarwal2020contextual}. For hybrid methods, we include \textbf{DBAL} \cite{zhdanov2019diverse}, which ranks samples based on the multiplication of their diversity and uncertainty scores. To mimic passive learning, we use \textbf{Random} selection, where each sample is assigned a score following a uniform distribution. To define the uncertainty function in NORIS, we use the least-confidence approach, where the uncertainty is determined by the predicted class probability. To provide a baseline for comparison, we report the results obtained from a "fully-trained" network trained on the entire training set.

\textbf{Results on PASCAL VOC.} Our experimental results on the PASCAL VOC dataset, presented in \cref{fig:pascal}, demonstrate the performance of our proposed selection strategy over the baseline methods. Our method outperforms all baselines by at least 1.0 mAP in the initial AL cycle. In the second cycle, our method outperforms the Random baseline by 3.4 mAP and the second-best method, CALD, by 1.5 mAP. As the number of actively selected samples grows, the two methods that combine uncertainty and diversity, NORIS and DBAL, consistently outperform the other baselines, highlighting the significance of utilizing uncertainty and diversity jointly in active learning. In the final cycle, where we train our detector with 90\% of the data, of which 60\% are actively sampled using our selection strategy, our proposed algorithm outperforms all baseline methods by at least 0.9 mAP. Notably, our approach achieves 90\% of the performance of a fully-trained model with only 60\% of the samples. This corresponds to 20\% less labeled data compared to the Random baseline and 5\% less labeled data compared to CALD. These results demonstrate that our proposed algorithm effectively selects informative samples while minimizing redundancy in the labeled dataset, resulting in improved data-efficiency and accuracy.

\textbf{Results on KITTI.} The performance of our proposed algorithm and the baseline methods on the KITTI dataset are presented in \cref{fig:kitti}. In the initial AL cycle, our method outperforms the Random baseline by 1.6 mAP and achieves a 1.0 mAP lead over the second-best method, CALD. Our approach maintains its lead over the baselines in subsequent cycles, and in the final cycle, where 60\% of the samples are actively selected, we outperform all the baselines by 0.8 mAP. Our proposed algorithm achieves 80\% of the fully-trained performance using only 50\% of the data, corresponding to a 30\% improvement in data savings compared to the Random baseline's 80\%.

\textbf{Results on CIFAR.} In CIFAR-10 (\cref{fig:cifar-10}), our method reaches the 80\% of the fully-trained performance using only 7.5\% of the data, with a data savings rate of 1.0\% compared to DBAL and 2.5\% compared to Random. In the final cycle, where we train with 20\% of the available data, our method reaches 1.0 higher accuracy. In CIFAR-100 (\cref{fig:cifar-100}), our method achieves 70\% of the fully-trained performance using only 12.5\% of the data, with a data savings rate of 1.0\% compared to DBAL and 5.0\% compared to Random. In the final cycle, where we train with 20\% of the available data, our method reaches 1.3 higher accuracy.

\subsection{Ablation on NORIS} \label{sec:exp_ablation}

\textbf{Variants of NORIS.} As we proposed two variants of NORIS - namely NORIS-Sum and NORIS-Max - along with two types of similarity scores (linear and Gaussian) in \cref{sec:method-redundancy}, we perform an ablation study to determine their relative performance. Our results, presented in \cref{table:ablation-variants}, indicate that NORIS-Sum tends to outperform NORIS-Max, and the Gaussian similarity score leads to superior performance over the linear similarity. We follow this configuration throughout our experiments.

\begin{table}
    \centering
    \begin{tabular}{|l|cc|}
    \hline
    Variant & Similarity & mAP \\
    \hline
    \multirow{2}{*}{NORIS-Max} & linear & 40.1 \\
                               & gaussian & \underline{40.8} \\
    \hline
    \multirow{2}{*}{NORIS-Sum} & linear & 40.4 \\
                               & gaussian & \textbf{41.0} \\
    \hline
    \end{tabular}
    \caption{Ablation study on variants of NORIS and similarity measures. Bold indicates the best-performing setting, and underline indicates the best-performing similarity measure within the NORIS-Max variant.}
    \label{table:ablation-variants}
\end{table}

\begin{figure*}
  \centering
  \begin{subfigure}{0.33\linewidth}
    \includegraphics[width=1.0\linewidth]{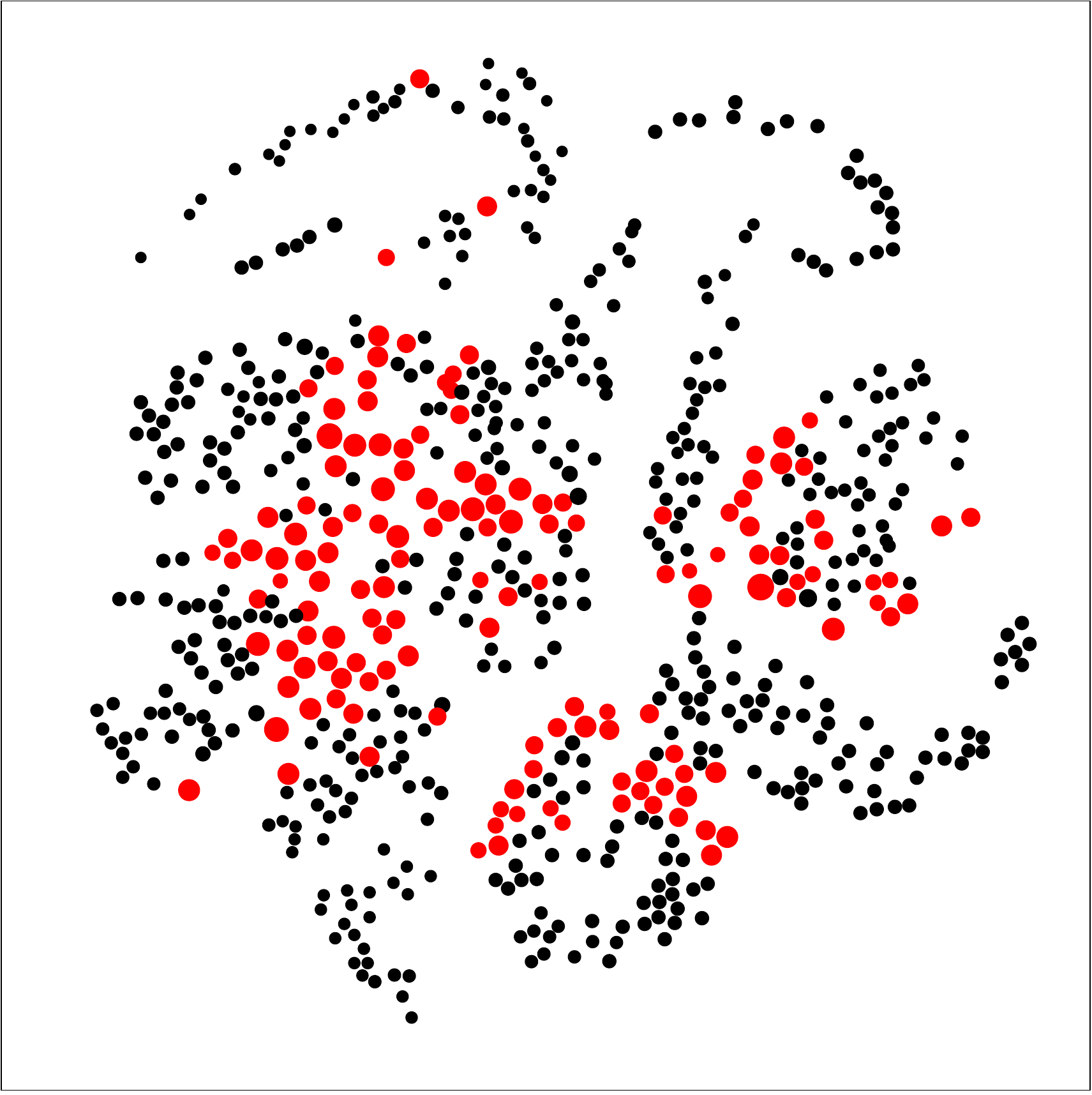}
    \caption{Uncertainty}
    \label{fig:vis-unc}
  \end{subfigure}
  \hfill
  \begin{subfigure}{0.33\linewidth}
    \includegraphics[width=1.0\linewidth]{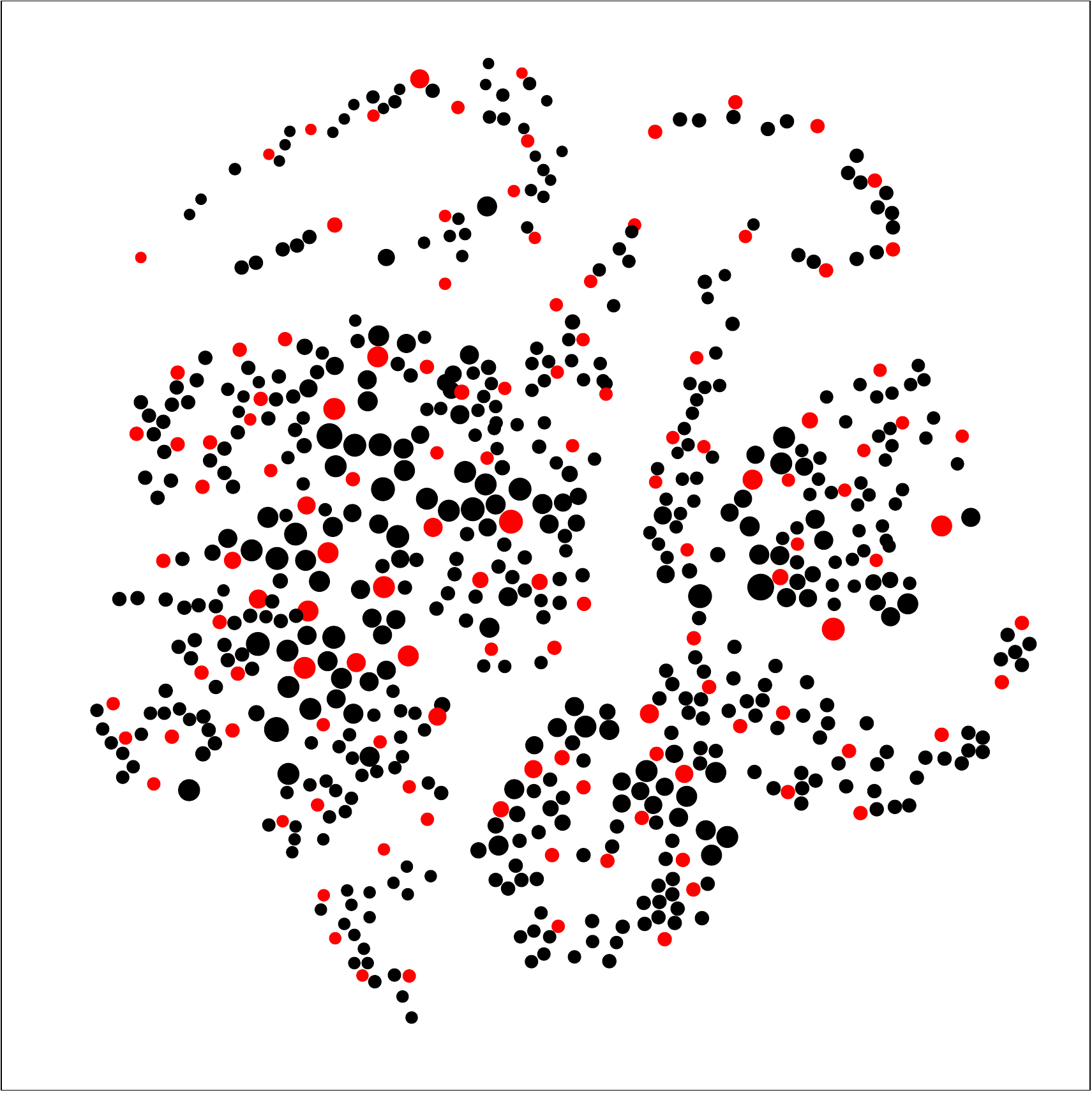}
    \caption{Diversity}
    \label{fig:vis-div}
  \end{subfigure}
  \hfill
  \begin{subfigure}{0.33\linewidth}
    \includegraphics[width=1.0\linewidth]{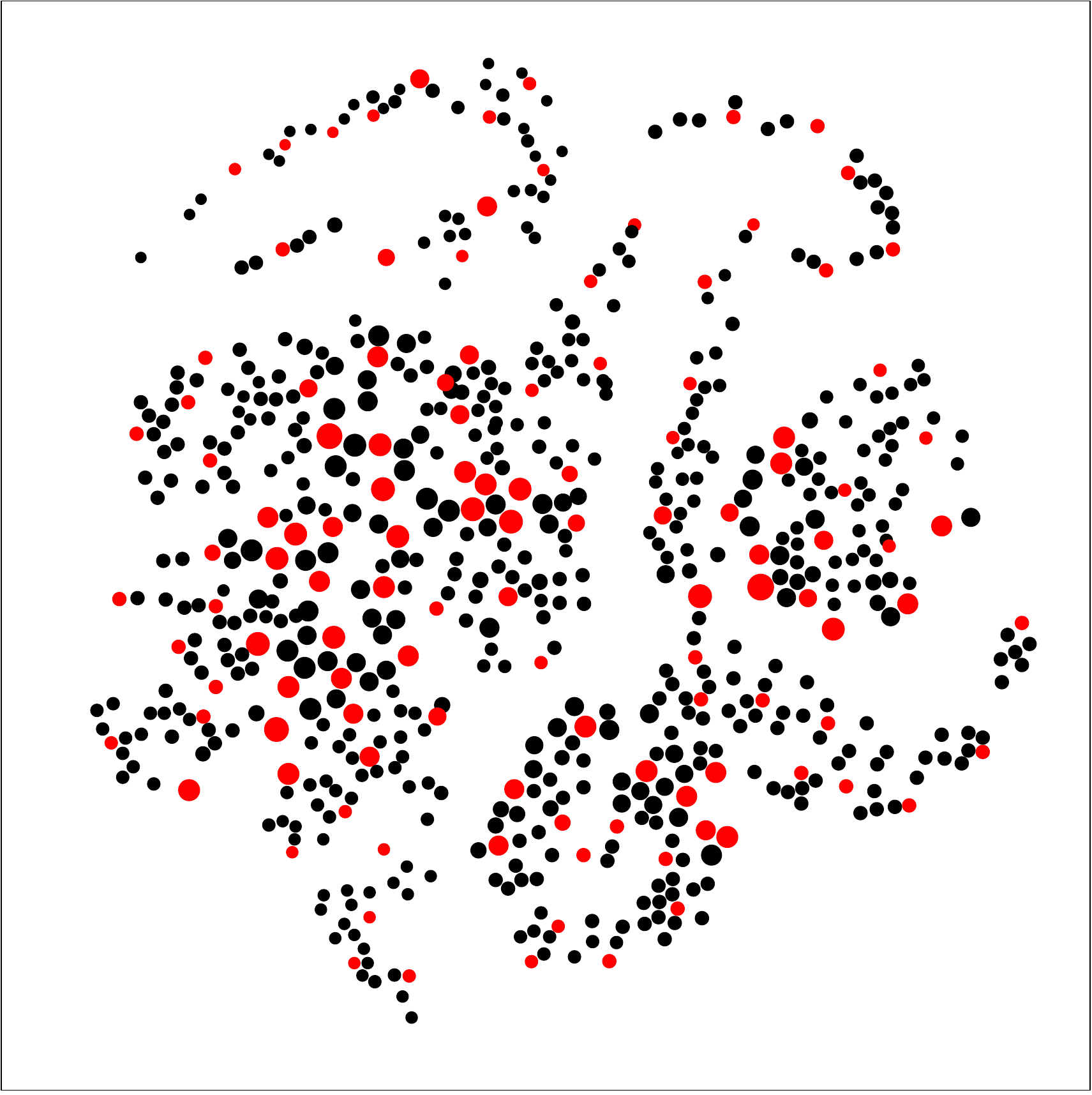}
    \caption{NORIS}
    \label{fig:vis-noris}
  \end{subfigure}
  \caption{t-SNE visualizations of feature embeddings for the CIFAR-10 dataset. Each dot corresponds to an image embedding, with its size being scaled in accordance with the uncertainty of the image. The red dots indicate the samples selected for labeling corresponding to the respective selection strategy.}
  \label{fig:tsne}
\end{figure*}

\textbf{t-SNE of selections.} To offer a visual perspective on how different selection methods operate, we provide a t-SNE visualization of selected points. The uncertainty-based method CALD, as shown in \cref{fig:vis-unc}, tends to select highly uncertain samples. However, it disregards samples situated in the outer regions, leading to the selection of numerous close-proximity samples, which introduced redundancy. CDAL, a diversity-based method illustrated in \cref{fig:vis-div}, selects more samples from the outer regions but tends to overlook some highly uncertain objects. NORIS, visualized in \cref{fig:vis-noris}, strikes a balance between selecting samples that are both informative and from outer regions.

\subsection{Ablation on Object Features} \label{sec:exp_ablation_od}

\textbf{Comparison of using image or object features.} To demonstrate the effectiveness of our proposal to use object features in contrast to image features for defining diversity, we adapt three SOTA diversity-based methods to work with object features: Core-Set\cite{sener2018active} VAAL\cite{sinha2019variational} and MAL\cite{ebrahimi2020minimax}. Specifically, we use features from the object detector and modify the discriminator heads of VAAL and MAL to output a similarity value between 0 and 1 for each detected object. We then compute a diversity score for each image by assigning the score of the discriminator output of its most dissimilar object. We refer the reader to the supplementary material for more detailed information on the modifications and a comparison of the layers from which the features are extracted within the CenterNet architecture.

We present our comparison results on the KITTI dataset in \cref{table:ablation_obj_vs_img}. Object features consistently outperform image features in all cycles for the MAL method. For Core-Set and VAAL, using image features initially leads to better performance, but using object features results in superior performance in later cycles. This observation can be attributed to the lower performance of the object detector during the initial cycles, making it more challenging to rely on the detected objects for diversity. Overall, the findings indicate that object features improve the performance of diversity-based AL algorithms compared to image features.

\begin{table}[htbp]
    \centering
    \begin{tabular}{|l|l|cccc|}
    \hline
     & & 40\% & 50\% & 70\% & 90\% \\
    \hline\hline
    \multirow{2}{*}{Core-set\cite{sener2018active}} & Img & \textbf{36.9} & \textbf{38.2} & 38.5 & 38.7 \\
     & Obj & 36.8 & 38.0 & \textbf{38.6} & \textbf{38.9} \\
    \hline
    \multirow{2}{*}{VAAL\cite{sinha2019variational}} & Img  & \textbf{36.6} & \textbf{38.2} & 38.5 & 39.1 \\
    & Obj & 36.3 & 38.1 & \textbf{39.0} & \textbf{39.4} \\
    \hline
    \multirow{2}{*}{MAL\cite{ebrahimi2020minimax}} & Img & 36.7 & 37.9 & 38.8 & 39.2 \\
    & Obj & \textbf{37.1} & \textbf{38.3} & \textbf{39.2} & \textbf{39.6} \\
    \hline
    \end{tabular}
    \caption{Comparison of using image or object features for three different diversity-based methods on KITTI. Bold values indicate the best performance for each method.}
    \label{table:ablation_obj_vs_img}
\end{table}

\begin{figure}
    \centering
    \includegraphics[width=0.99\linewidth]{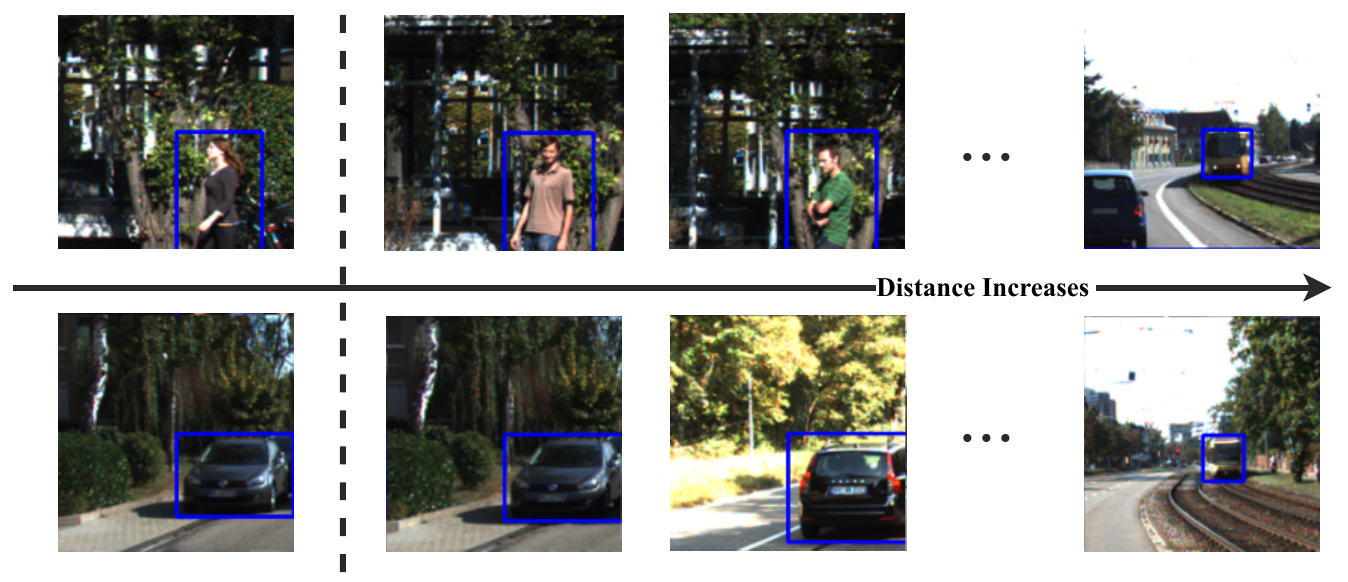}
    \caption{Visualization of the distances computed using object features. The leftmost column corresponds to the selected query objects. For each query object, the available objects from the unlabeled dataset are ranked in ascending order of their distance to the respective query sample from left to right.}
    \label{fig:ablation_visual}
\end{figure}

\textbf{Visual comparison.} In \cref{fig:ablation_visual}, we compare qualitatively the selection strategy of Core-Set with object features. Our analysis reveals that the query objects exhibit similar characteristics to their most similar object. Objects belong to the same class and have similar visual appearances, like the color of the vehicles in the second row. This suggests that using object features can identify these patterns and select dissimilar objects, resulting in a more diverse and representative training set. 

\section{Conclusion} \label{sec:conclusion}

We introduced NORIS, a novel AL strategy that combines uncertainty with diversity to select the most informative and non-redundant samples for object detection. Our approach jointly optimizes both criteria by reducing the information score of a sample if it is close to other selected samples with high uncertainty. To better define diversity for object detection, we proposed using the distance between object features to measure the similarity of samples. Our experiments demonstrated that incorporating non-redundant sampling with object feature-based diversity led to better performance than SOTA AL methods. Moreover, we observed improved performance for all diversity-based methods when utilizing object features.

{\small
\bibliographystyle{ieee_fullname}
\bibliography{references}
}

\clearpage

\twocolumn[{
\renewcommand\twocolumn[1][]{#1}
\centering
\Large
\textbf{Active Learning for Object Detection with Non-Redundant Informative Sampling} \\
\vspace{0.5em}Supplementary Material \\
\vspace{1.0em}

}]

\setcounter{section}{0}

\section{Loss Inequality \cref{eq:loss_bound}}\label{sec:supp_loss_inequality}

We denote the losses of a sample $u$ before and after the training with a sample $v$ as $l(u;\theta_{\text{i}})$ and $l(u;\theta_{\text{i+1}})$, and similarly for $v$. The $\kappa$-Lipschitz continuity of the loss function $l(\cdot)$ (we refer to the Core-Set\cite{sener2018active} for proof) provides the following inequality for any points $u$ and $v$,
\begin{equation}
    |l(u;\theta) - l(v;\theta)| \leq \kappa * d(u, v)
\end{equation}

Before the training of $v$, we have the inequality,
\begin{equation}
    |l(u; \theta_{\text{i}}) - l(v; \theta_{\text{i}})| \leq \kappa * d(u, v)
\end{equation}

Since we select the sample with a higher loss, we assume that the initial loss at $v$ is higher than the initial loss at $u$,
\begin{equation}
\label{eq:supp_coreset_init}
    l(v; \theta_{\text{i}}) - l(u; \theta_{\text{i}}) \leq \kappa * d(u, v)
\end{equation}

Similarly, for the final weights $\theta_{\text{i+1}}$, 
\begin{equation}
    |l(u; \theta_{\text{i+1}}) - l(v; \theta_{\text{i+1}})| \leq \kappa * d(u, v)
\end{equation}

Following Core-Set\cite{sener2018active}, we assume the final loss at $v$, $l(v;\theta_{\text{i+1}})$, to be minimized, approaching zero. Therefore, after the training of $v$, we have the inequality,
\begin{equation}
\label{eq:supp_coreset_final}
    l(u; \theta_{\text{i+1}}) \leq \kappa * d(u, v)
\end{equation}

Summing up the two inequalities in \cref{eq:supp_coreset_init}, \cref{eq:supp_coreset_final},
\begin{equation}
    l(v; \theta_{\text{i}}) - l(u; \theta_{\text{i}})+l(u; \theta_{\text{i+1}}) \leq 2 * \kappa * d(u, v)
\end{equation}

Isolating $l(u; \theta_{\text{i+1}})$,
\begin{equation}
    l(u; \theta_{\text{i+1}}) \leq l(u; \theta_{\text{i}}) + 2 * \kappa * d(u, v) - l(v; \theta_{\text{i}})
\end{equation}

\begin{table*}[t]
    \centering
    \begin{tabular}{|l|c|ccccccccccc|}
    \hline
    \multirow{2}{*}{Variant} & \multirow{2}{*}{Similarity} & \multicolumn{11}{c|}{$\alpha$} \\
    \cline{3-13}
    &  & 0.01 & 0.1  & 0.2  & 0.3  & 0.4  & 0.5  & 0.6  & 0.7  & 0.8  & 0.9  & 1.0 \\
    \hline\hline
    \multirow{2}{*}{NORIS-Max} & linear & 39.4 & 39.7 & 39.6 & 39.8 & 39.7 & 39.0 & 39.8 & 39.7 & \textbf{40.1} & 39.5 & 40.0 \\
                               & gaussian & 40.6 & 40.4 & \textbf{40.8} & 40.3 & 40.5 & 40.5 & 39.9 & 40.2 & 40.3 & 40.3 & 39.7 \\
    \hline
    \multirow{2}{*}{NORIS-Sum} & linear & 39.6 & 39.5 & 39.1 & 39.0 & 39.5 & 39.8 & 40.2 & 39.9 & 39.5 & 40.1 & \textbf{40.4} \\
                               & gaussian & 40.0 & 40.4 & 40.2 & 40.5 & 40.6 & \textbf{41.0} & 40.3 & 40.6 & 40.4 & 40.0 & 40.7 \\
    \hline
    \end{tabular}
    \caption{Hyperparameter search for two variants of NORIS and the two similarity measures. Results are reported on the KITTI dataset with the mAP metric. Bold indicates the best-performing parameter value for the row.}
    \label{table:ablation-hyperparameter}
\end{table*}

\section{Hyperparameter $\lambda$}

As stated in the main manuscript, we use $\lambda = \alpha \cdot d_{\text{max}}$ for linear similarity and $\lambda = \frac{(\alpha \cdot d_{\text{max}})^2}{\pi}$ for Gaussian similarity. In this section, we explain the reasoning behind our choice of defining the Gaussian similarity parameter in this manner.

Let $\lambda_G$ and $\lambda_l$ denote the hyperparameter for the Gaussian and linear similarity function respectively. Our goal is to choose $\lambda_G$ such that the Gaussian and the linear similarity function have a similar behavior. We argue that this happens when the integral between the two functions is 0:

\begin{equation}
    \int\limits_0^\infty e^{- \frac{1}{\lambda_G} s^2} - \max\left\{0, 1 - \frac{s}{\lambda_l}\right\} ds \overset{!}{=} 0
\end{equation}

This is equivalent to require:
\begin{equation}
\label{eq: gaussian equal linear integral}
    \int\limits_0^\infty e^{- \frac{1}{\lambda_G} s^2} ds \overset{!}{=} \int\limits_0^\infty \max\left\{0, 1 - \frac{s}{\lambda_l}\right\} ds
\end{equation}
Let's calculate both sides. We know for a fact that the integral over the real numbers of the probability density function of a Gaussian distribution is 1:
\begin{equation*}
    \frac{1}{\sigma_G \sqrt{2 \pi}} \int_\mathbbm{R} e^{- \frac{s^2}{2 \sigma_G^2}} ds = 1 \text{ for all } \sigma_G > 0
\end{equation*}
By the change of variable $\lambda_G = 2 \sigma_G^2$, we obtain
\begin{equation*}
    \int_\mathbbm{R} e^{- \frac{s^2}{\lambda_G}} ds = \sqrt{\lambda_G \pi} \text{ and thus} \int\limits_0^\infty e^{- \frac{s^2}{\lambda_G}} ds = \frac{\sqrt{\lambda_G \pi}}{2}
\end{equation*}
for the left hand side of \cref{eq: gaussian equal linear integral} by the symmetry. Next, we simplify the right hand side:
\begin{equation*}
    \int\limits_0^\infty \max\left\{0, 1 - \frac{s}{\lambda_l}\right\} ds = \int\limits_0^{\lambda_l} 1 - \frac{s}{\lambda_l} ds = \left(s - \frac{s^2}{2 \lambda_l}\right) \mid_0^{\lambda_l} = \frac{\lambda_l}{2}
\end{equation*}
From the calculations, we infer that the integrals in \cref{eq: gaussian equal linear integral} are equal when $\frac{\sqrt{\lambda_G \pi}}{2} = \frac{\lambda_l}{2}$ which can be achieved by setting
\begin{equation}
\label{eq: infer lambda from linear to gaussian}
    \lambda_G = \frac{\lambda_l^2}{\pi}.
\end{equation}
Intuitively, this implies that the Gaussian similarity score is $e^{- \pi} \approx 0.043$ at a distance of $\lambda_l$. When using the Gaussian similarity, we therefore suggest fixing a value $\alpha \in (0, 1]$ and scaling to $\lambda = \frac{(\alpha \cdot d_{\text{max}})^2}{\pi}$ in each AL cycle for hyperparameter tuning. An advantage of this theoretical investigation is that once a good hyperparameter choice $\lambda_l > 0$ for the linear similarity function is found, we can act on the assumption that setting $\lambda_G = \frac{\lambda_l^2}{\pi}$ in the Gaussian similarity function performs similar and vice versa.

\textbf{Hyperparameter optimization for $\alpha$.} As explained in the manuscript, we select a fixed value for $\alpha$ and scale it with $d_{\text{max}}$ to derive the corresponding $\lambda$ value. Thus, determining the optimal value of $\alpha$ is critical in our work. We present the results of our parameter search in \cref{table:ablation-hyperparameter}. Notably, we observe that higher values of $\alpha$ (0.8 and 1.0) tend to yield superior performance for the linear similarity. Conversely, for the Gaussian similarity, the most effective values are 0.2 and 0.5 for the NORIS-Max and NORIS-Sum variants, respectively.

\section{NORIS-Max Algorithm}

We present the NORIS-Max algorithm in \cref{alg:noris-max}. In each iteration of the while-loop, the algorithm selects the sample $u_{next}$ that leads to the highest immediate gain and adds it to $S$. Subsequently, for the most similar sample $u_c = \argmax_{u \in X_{\text{U}} \setminus S}(sim(u,u_{next}))$, the information score is updated by subtracting the uncertainty of $u_{\text{next}}$ scaled by the similarity $\text{sim}(u, u_{\text{next}})$. The runtime complexity of this algorithm is $\mathcal{O}(B \cdot \lvert X_{\text{U}} \rvert)$.

\begin{algorithm}
    \caption{NORIS-Max}
    \label{alg:noris-max}
    \begin{algorithmic}
        \Require unlabeled dataset $X_{\text{U}}$, batch size $1 \leq B \leq \abs{X_{\text{U}}}$ 
        \State $S \gets \emptyset$
        \While{$\abs{S} \neq B$}
\State $u_{\text{next}} \gets \argmax\limits_{u \in X_{\text{U}} \setminus S} \sigma(u)$
            \State $S \gets S \cup \{u_{\text{next}}\}$
            \State $u_c \gets \argmax_{u \in X_{\text{U}} \setminus S}(sim(u,u_{next}))$
            \State $\sigma(u_c) \gets \sigma(u_c) - \text{sim}(u_c, u_{next}) * \sigma(u_{next})$
        \EndWhile
    \end{algorithmic}
\end{algorithm}

\section{Extending MAL and VAAL to Object Features}\label{sec:mal_vaal}

\begin{figure}[htbp]
    \centering
    \includegraphics[width=0.98\linewidth]{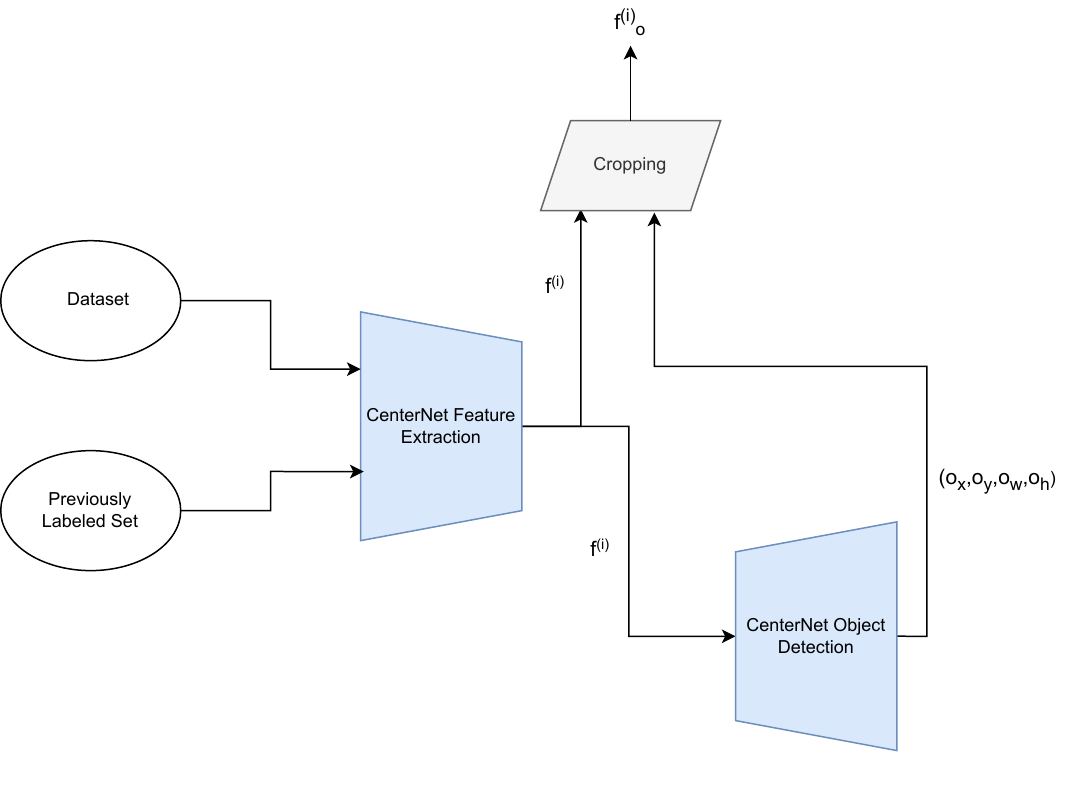}
    \caption{Illustration of extracting object features from the CenterNet detector.}
    \label{fig:vmp1}
\end{figure}

We provide a comprehensive description of how we incorporate object features into the VAAL \cite{sinha2019variational} and MAL \cite{ebrahimi2020minimax} methods. \Cref{fig:vmp1} demonstrates the process of extracting object features from the CenterNet detector. Following object detection, we crop the corresponding bounding box locations from the intermediate feature map to obtain object features, as described in the main manuscript.

\begin{figure}
\begin{center}
  \centering
  \begin{subfigure}[t]{0.34\linewidth}
    \includegraphics[width=1.0\linewidth]{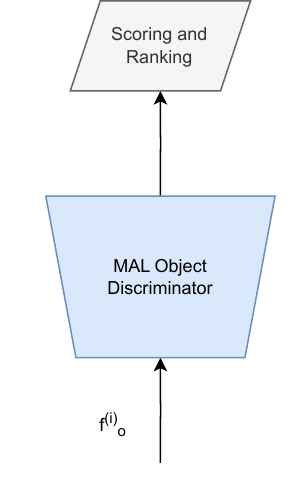}
    \caption{MAL.}
    \label{fig:vmp2-mal}
  \end{subfigure}
  \begin{subfigure}[t]{0.64\linewidth}
    \includegraphics[width=1.0\linewidth]{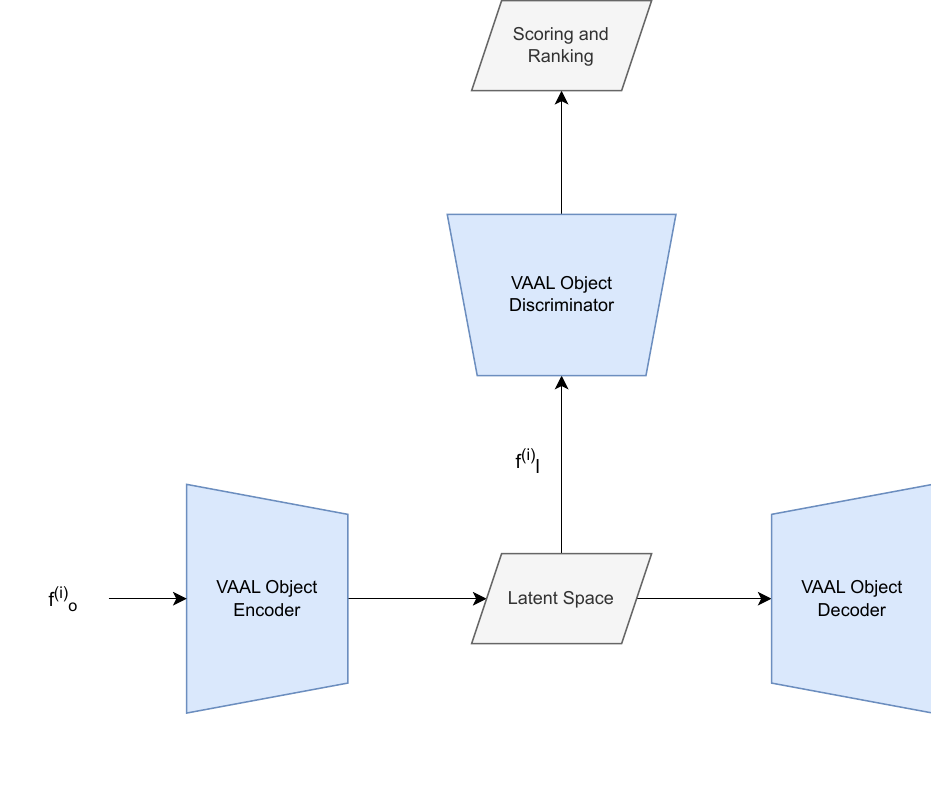}
    \caption{VAAL utilizes VAE.}
    \label{fig:vmp2-vaal}
  \end{subfigure}
  \caption{Illustration of the VAAL and MAL with object features.}
  \label{fig:vmp2}
\end{center}
\end{figure}

\cref{fig:vmp2-mal} and \cref{fig:vmp2-vaal} depict the utilization of object features for the MAL and VAAL methods, respectively. For the MAL approach, we input the object features into the discriminator, and for the VAAL method, we feed the object features to a variational autoencoder (VAE). In both cases, we employ the discriminator predictions as a similarity score to the labeled set and rank images based on the most dissimilar object it contains.

\section{CenterNet Features}\label{sec:centernet_features}

\begin{figure}[htbp]
    \centering
    \includegraphics[width=0.9\linewidth]{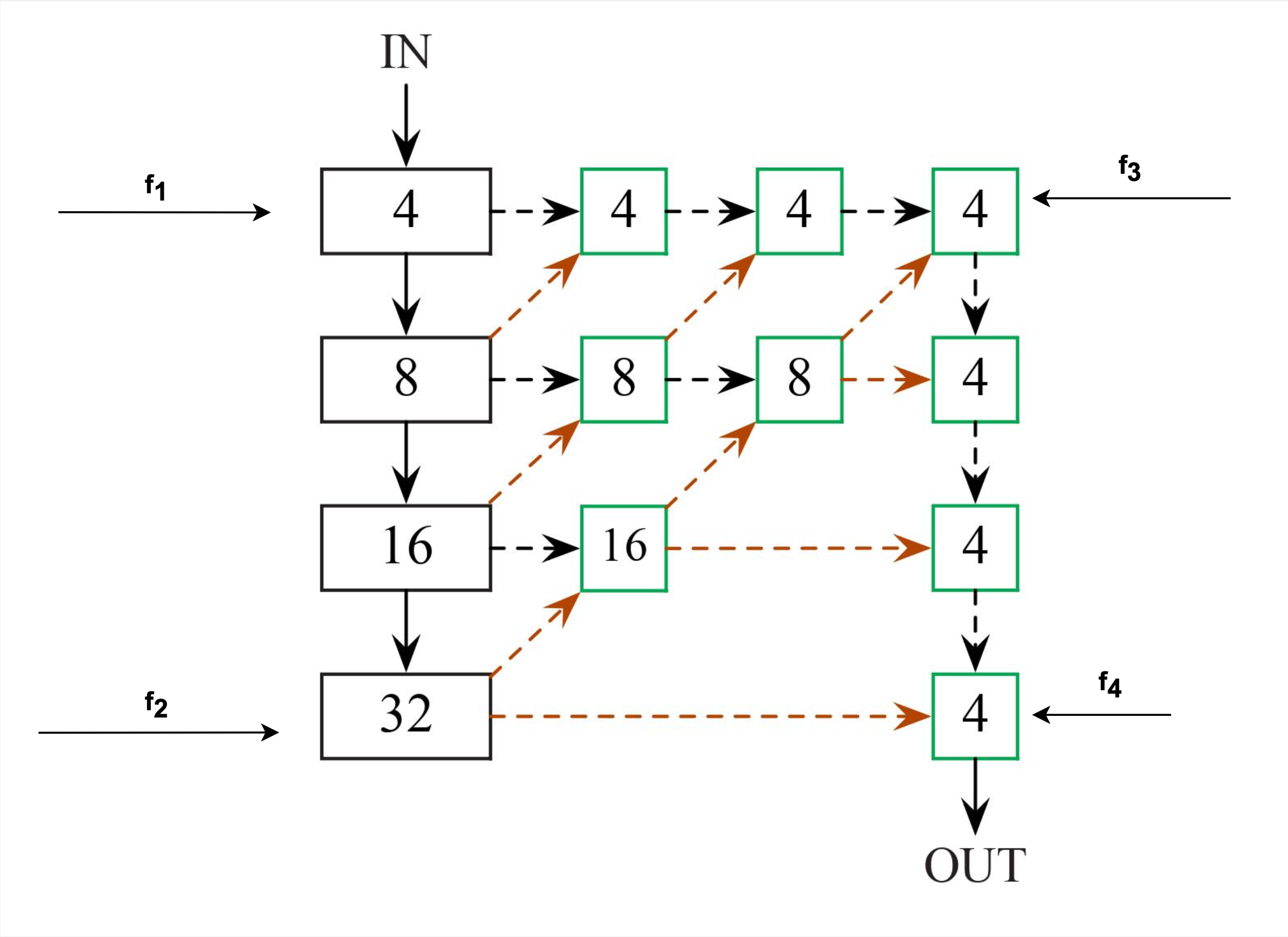}
    \caption{Illustration showcasing the intermediate network layers employed in our experiments. The object features are extracted from the layers represented in this visualization.}
    \label{fig:cwf}
\end{figure}

In \cref{fig:overview}, we describe the process of extracting object features. An important consideration is determining which layer from the detector represents the intermediate features. In \cref{fig:cwf}, we provide a detailed illustration of different intermediate features from the CenterNet detector \cite{zhou2019objects}. We denote the features extracted from the detector as f1, f2, f3, f4, with f1 being the closest feature to the input image and f4 being the feature closest to the output as depicted in \cref{fig:cwf}.

\begin{figure}
    \includegraphics[width=1.0\linewidth]{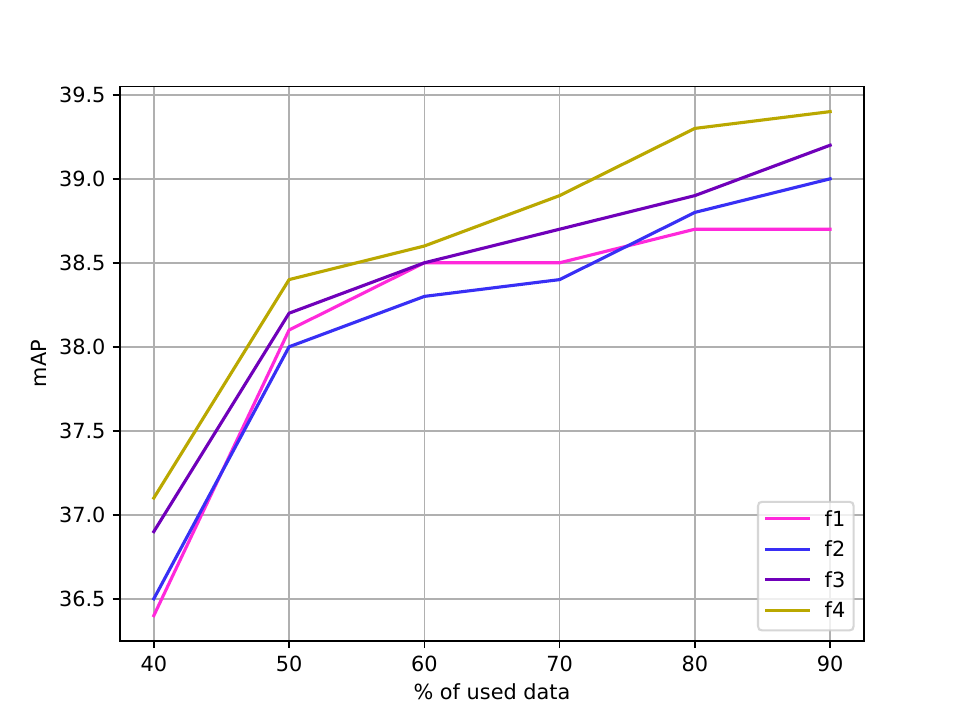}
    \caption{Comparison of using feature maps from different layers of the detector on KITTI. f1 represent the closest feature to the input and f4 represent the furthest.}
    \label{fig:ablation_features}
\end{figure}

Then, in \cref{fig:ablation_features}, we perform an ablation study to investigate the impact of using different intermediate feature maps from various layers of the CenterNet detector in our selection strategy. The results indicate that using features from the later layers leads to improved performance in our selection strategy. This finding indicates that utilizing task-specific features can be advantageous.

\section{Ablation of the distance metric, aggregation function, and use of image features.} To investigate the impact of different design choices on the performance of our approach, we compare using different distance metrics, aggregation functions, and features. Specifically, we compare using cosine or Euclidean distance as the distance metric, aggregating object features into an image score using averaging or maximum, and using image features or not. Our results in \cref{table:ablation-distance} indicate that using image features leads to higher performance than not using them. The choice of the aggregation function has a negligible impact on the outcome, but we observe a slight improvement when using the maximum over averaging. Our results show a slight advantage of using Euclidean distance over cosine distance as the distance metric. We choose to use Euclidean distance, maximum aggregation and include image features for our experiments, as this configuration corresponds to the last row with the highest performance in our results.

\begin{table}[b]
    \centering
    \begin{tabular}{|l|l|l|cccc|}
    \hline
    Dist. & Agg. & $f_I$ & 40\% & 60\% & 80\% & 100\% \\
    \hline\hline
    cos & avg &  & 36.6 & 38.3 & 38.6 & 38.6 \\
    euc & avg &  & 37.0 & 38.1 & 38.5 & 39.1 \\
    cos & max &  & 36.7 & 38.1 & 38.6 & 39.0 \\
    euc & max &  & 36.4 & 38.0 & 39.0 & 39.3 \\
    cos & avg & \checkmark & 36.9 & 37.9 & 38.7 & 39.4 \\
    euc & avg & \checkmark & 37.0 & 38.5 & 39.2 & 39.5 \\
    cos & max & \checkmark & 37.0 & 38.3 & 39.0 & 39.4 \\
    euc & max & \checkmark & 37.2 & 38.2 & 39.4 & 39.6 \\
    \hline
    \end{tabular}
    \caption{Comparison of distance metric, aggregation function and using image features on KITTI.}
    \label{table:ablation-distance}
\end{table}

\section{Exact Values from AL Figures and Variances}

Due to space constraints in the main paper, we present our AL comparisons as plots. However, in the supplementary material, we provide the precise values and variances for our experiments. \cref{table:supp-fig3a}, \cref{table:supp-fig3b}, \cref{table:supp-fig3c} and \cref{table:supp-fig3d} provides the exact metric values for Figures \ref{fig:pascal}, \ref{fig:kitti}, \ref{fig:cifar-10} and \ref{fig:cifar-100} from the main paper. The mean and variances of three experiments trained with different random initializations are presented.

\begin{table*}
    \centering
    \begin{tabular}{|l|cccccc|}
    \hline
     & 40 & 50 & 60 & 70 & 80 & 90 \\
    \hline\hline
    Random & 60.80$\pm$0.28 & 64.10$\pm$0.27 & 66.10$\pm$0.27 & 68.10$\pm$0.18 & 68.80$\pm$0.37 & 69.20$\pm$0.23 \\
    CDAL & 62.40$\pm$0.32 & 64.80$\pm$0.34 & 67.80$\pm$0.16 & 68.80$\pm$0.36 & 69.80$\pm$0.21 & 70.50$\pm$0.23 \\
    LL4AL & 60.60$\pm$0.35 & 64.40$\pm$0.28 & 66.70$\pm$0.16 & 68.90$\pm$0.24 & 69.90$\pm$0.25 & 70.80$\pm$0.29 \\ 
    DBAL & 61.60$\pm$0.32 & 65.20$\pm$0.22 & 67.70$\pm$0.19 & 69.50$\pm$0.37 & 71.10$\pm$0.29 & 71.50$\pm$0.29 \\
    CALD & 62.90$\pm$0.16 & 66.00$\pm$0.34 & 67.80$\pm$0.18 & 69.10$\pm$0.33 & 70.40$\pm$0.35 & 71.20$\pm$0.16 \\
    NORIS & 63.90$\pm$0.19 & 67.50$\pm$0.26 & 68.50$\pm$0.32 & 70.10$\pm$0.36 & 71.60$\pm$0.38 & 72.40$\pm$0.20 \\
    \hline
    \end{tabular}
    \caption{Comparison of mAP with SOTA AL methods on PASCAL-VOC 2007. (\cref{fig:pascal})}
    \label{table:supp-fig3a}
\end{table*}

\begin{table*}
    \centering
    \begin{tabular}{|l|cccccc|}
    \hline
     & 40 & 50 & 60 & 70 & 80 & 90 \\
    \hline\hline
    Random & 36.10$\pm$0.17 & 37.30$\pm$0.11 & 37.90$\pm$0.06 & 38.40$\pm$0.06 & 38.70$\pm$0.11 & 38.80$\pm$0.16 \\
    CALD & 36.90$\pm$0.10 & 38.20$\pm$0.08 & 38.60$\pm$0.12 & 38.90$\pm$0.15 & 39.10$\pm$0.16 & 39.20$\pm$0.20 \\
    LL4AL & 35.60$\pm$0.08 & 37.70$\pm$0.07 & 38.50$\pm$0.10 & 38.80$\pm$0.14 & 39.30$\pm$0.12 & 39.40$\pm$0.16 \\ 
    DBAL & 36.00$\pm$0.12 & 37.80$\pm$0.17 & 38.60$\pm$0.14 & 39.60$\pm$0.06 & 39.80$\pm$0.13 & 40.00$\pm$0.06 \\
    CALD & 36.70$\pm$0.10 & 38.50$\pm$0.05 & 39.20$\pm$0.16 & 39.40$\pm$0.11 & 39.50$\pm$0.10 & 39.60$\pm$0.14 \\
    NORIS & 37.70$\pm$0.08 & 38.80$\pm$0.13 & 39.50$\pm$0.13 & 39.70$\pm$0.12 & 40.20$\pm$0.10 & 40.80$\pm$0.12 \\
    \hline
    \end{tabular}
    \caption{Comparison of mAP with SOTA AL methods on KITTI \textit{val}. (\cref{fig:kitti})}
    \label{table:supp-fig3b}
\end{table*}

\begin{table*}
    \centering
    \begin{tabular}{|l|ccccccc|}
    \hline
     & 4.0 & 6.0 & 8.0 & 10.0 & 12.0 & 14.0 & 16.0 \\
    \hline\hline
    Random & 59.30$\pm$0.54 & 63.40$\pm$0.33 & 67.00$\pm$0.75 & 70.60$\pm$0.56 & 71.70$\pm$0.58 & 73.00$\pm$0.70 & 74.20$\pm$0.69 \\
    CDAL & 58.40$\pm$0.75 & 63.60$\pm$0.45 & 67.90$\pm$0.34 & 69.80$\pm$0.65 & 72.50$\pm$0.45 & 73.50$\pm$0.31 & 75.50$\pm$0.39 \\
    LL4AL & 59.10$\pm$0.77 & 64.90$\pm$0.71 & 69.30$\pm$0.71 & 70.90$\pm$0.76 & 73.30$\pm$0.70 & 74.30$\pm$0.74 & 76.00$\pm$0.44 \\
    DBAL & 59.30$\pm$0.67 & 65.20$\pm$0.76 & 68.40$\pm$0.32 & 73.10$\pm$0.69 & 74.30$\pm$0.32 & 76.70$\pm$0.35 & 77.30$\pm$0.45 \\
    LC & 58.90$\pm$0.31 & 64.80$\pm$0.48 & 68.80$\pm$0.65 & 71.50$\pm$0.33 & 73.30$\pm$0.65 & 74.90$\pm$0.39 & 77.50$\pm$0.73 \\
    NORIS & 61.20$\pm$0.55 & 66.80$\pm$0.69 & 70.60$\pm$0.71 & 74.20$\pm$0.66 & 75.80$\pm$0.69 & 76.90$\pm$0.44 & 78.80$\pm$0.49 \\
    \hline
    \end{tabular}
    \caption{Comparison of mAP with SOTA AL methods on CIFAR-10. (\cref{fig:cifar-10})}
    \label{table:supp-fig3c}
\end{table*}

\begin{table*}
    \centering
    \begin{tabular}{|l|ccccccc|}
    \hline
     & 4.0 & 6.0 & 8.0 & 10.0 & 12.0 & 14.0 & 16.0 \\
    \hline\hline
    Random & 41.90$\pm$0.33 & 43.10$\pm$0.17 & 45.70$\pm$0.13 & 46.90$\pm$0.37 & 47.90$\pm$0.31 & 48.40$\pm$0.24 & 49.40$\pm$0.10 \\
    CDAL & 41.60$\pm$0.16 & 43.50$\pm$0.28 & 45.80$\pm$0.26 & 47.50$\pm$0.14 & 48.80$\pm$0.38 & 49.40$\pm$0.12 & 49.90$\pm$0.34 \\
    LL4AL & 41.60$\pm$0.24 & 43.90$\pm$0.27 & 46.50$\pm$0.37 & 48.00$\pm$0.10 & 48.90$\pm$0.14 & 49.00$\pm$0.25 & 49.80$\pm$0.40 \\
    DBAL & 41.10$\pm$0.10 & 44.90$\pm$0.33 & 47.20$\pm$0.25 & 47.90$\pm$0.38 & 49.10$\pm$0.14 & 50.10$\pm$0.10 & 50.80$\pm$0.14 \\
    LC & 41.70$\pm$0.19 & 44.40$\pm$0.10 & 45.80$\pm$0.18 & 47.80$\pm$0.12 & 48.60$\pm$0.20 & 49.40$\pm$0.23 & 50.00$\pm$0.10 \\
    NORIS & 41.90$\pm$0.15 & 45.50$\pm$0.25 & 47.70$\pm$0.38 & 48.30$\pm$0.40 & 49.50$\pm$0.13 & 50.70$\pm$0.32 & 51.00$\pm$0.16 \\
    \hline
    \end{tabular}
    \caption{Comparison of mAP with SOTA AL methods on CIFAR-100. (\cref{fig:cifar-100})}
    \label{table:supp-fig3d}
\end{table*}

\end{document}